# Deliberation and its Role in the Formation of Intentions[*]

**Anand S. Rao**
Australian Artificial Intelligence Institute
Carlton, Victoria 3053
Australia
Email: anand@aaii.oz.au

**Michael P. Georgeff**
Australian Artificial Intelligence Institute
Carlton, Victoria 3053
Australia
Email: georgeff@aaii.oz.au

## Abstract

Deliberation plays an important role in the design of rational agents embedded in the real-world. In particular, deliberation leads to the formation of intentions, i.e., plans of action that the agent is committed to achieving. In this paper, we present a branching-time possible-worlds model for representing and reasoning about, beliefs, goals, intentions, time, actions, probabilities, and payoffs. We compare this possible-worlds approach with the more traditional decision-tree representation and provide a transformation from decision trees to possible worlds. Finally, we illustrate how an agent can perform deliberation using a decision-tree representation and then use a possible-worlds model to form and reason about his intentions.

## 1 INTRODUCTION

The design of rational agents, situated in a dynamic world and operating effectively under real-time constraints and resource limitations, has been of great interest to researchers in philosophy, artificial intelligence, and computer science [1, 2, 7]. Such rational agents have to balance the time taken thinking against the time needed for acting. In particular, they must balance the frequency of reassessment of options against continuing commitment to previously chosen plans.

Classical planning addresses only one aspect of the above problem; namely, means-end reasoning. Means-end reasoning involves finding a sequence of actions that satisfy a certain end goal (or goals). However, simplifying assumptions are made about the capabilities of the reasoning agents and the worlds they occupy that limit the use of these techniques to essentially static domains.

Classical decision theory, on the other hand, addresses the problem of weighing alternative courses of action and choosing the best plan of action according to some well-defined criteria, such as maximizing expected utility. However, this theory presupposes an ideal agent who can consider and weigh all possible alternative courses of action before making a decision. In real situations, such an assumption is rarely valid—not only does the world undergo continuous change, even as the agent is deliberating, but the agent may not be capable of enumerating all the alternatives.

What is required for the design of rational agents is a combination of symbolic means-end reasoning and numeric decision-theoretic analysis that takes into account the resource-boundedness of rational agents [6]. One such design is provided by the belief-desire-intention (BDI) architecture [2]. This architecture gives primary importance to the attitude of intentions. While most philosophical accounts of rational agency treat intentions as being reducible to beliefs and desires, Bratman [1] argues convincingly that intentions, especially future-directed intentions, play a significant and distinct role in resource-bounded reasoning.

Bratman treats intentions as plans of action that the agent is committed to achieving. Prior intentions constrain the search for possible means for achieving the current intention and thus focus the means-end reasoning process. The notion of commitment, which lends a certain sense of stability to means-end reasoning, is balanced against the notion of reconsideration of intentions, which lends a certain sense of reactivity. This rational balance between commitment and reconsideration is essential for effective means-end reasoning in dynamic domains.[1]

---

[*]This research was in part supported by a *Generic Industry Research and Development Grant* from the Department of Industry, Technology and Commerce, Australia and in part by the Australian Civil Aviation Authority.

[1]Some interesting experimental work has recently been done in this area [11].

Intentions play two important roles in resource-bounded decision-theoretic analysis or deliberation [1]. First, prior intentions pose problems for further deliberation, i.e., prior intentions produce the decision problems that the agent needs to consider. Second, prior intentions constrain the deliberation process because they rule out options that conflict with existing intentions. Under this view, the deliberation process is a continuous resource-bounded activity rather than a one-off exhaustive decision-theoretic analysis.

So far, we have discussed the role of intentions in means-ends reasoning and deliberation. However, we have not discussed how the agent arrives at his intentions. Prior intentions, means-ends reasoning, and deliberation are all involved in the formation of intentions. By means-ends reasoning, a prior intention towards an end results in the agent enumerating all the alternative or means of achieving this end; the agent by deliberating on all these alternatives then chooses the best one and commits to it by forming an intention.

We have previously provided [14, 15] a logical framework that describes the role of intentions in means-end reasoning. In this paper, we illustrate how the process of deliberation can lead to the formation of intentions.

## 2 OVERVIEW

BDI-architectures are formalized by defining notions such as beliefs, goals, intentions, actions, and the inter-relationships between them. We have previously shown how this can be accomplished using a branching-time possible-worlds logic [15].

Briefly, the structure of our logic is as follows: Each world is a temporal structure with a branching time future and a single past called a *time tree* [4]. A particular time point in a particular world is called a *situation*. Event types transform one time point into another. For each situation we associate a set of *belief-accessible*, *goal-accessible*, and *intention-accessible* worlds; intuitively, those worlds that the agent *believes* to be possible, *desires* to bring about, and *commits* to achieving, respectively. Multiple possible worlds result from the agent's lack of knowledge about the state of the world. But within each of these possible worlds, the branching future represents the *choice* of actions available to the agent. Moving from belief to goal to intention worlds amounts to successively pruning the paths of the time tree; intuitively, to making increasingly selective choices about one's future actions. This is captured semantically by requiring that for each belief-accessible world there exists a sub-world which is goal-accessible and, in turn, for each goal-accessible world there exists a sub-world which is intention-accessible (see Figure 1).

In this paper, we extend the expressive power of the above logic to model the process of deliberation by introducing *subjective probabilities* and *subjective payoffs*. For the former, we adopt the formalism of Fagin and Halpern [5] and extend it to a branching-time model. For the latter, we introduce a payoff function that associates numeric values (or payoffs) with certain paths in a time tree. Intuitively, an agent at each situation has a probability distribution on his belief-accessible worlds. He then chooses sub-worlds of these that he considers are worth pursuing and associates a payoff value with each path in these sub-worlds. These sub-worlds are considered to be the agent's goal-accessible worlds. By making use of the probability distribution on his belief-accessible worlds and the payoff distribution on the paths in his goal-accessible worlds, the agent determines the best plan(s) of action for different scenarios. This process will be called *Possible-Worlds(PW) deliberation*. The result of PW-deliberation is a set of sub-worlds of the goal-accessible worlds; namely, the ones that the agent considers best. These sub-worlds are taken to be the intention-accessible worlds that the agent *commits* to achieving.

In contrast to this approach, decision theory represents the problem as a decision tree (or, equivalently, as a payoff matrix or influence diagram). A decision tree consists of three types of nodes: (a) *decision nodes*, which represent the choice of actions; (b) *chance nodes*, which represent the state of uncertainty in the world; and (c) *terminal nodes*, which represent the value of outcomes. Based on the category of decision making – namely, certainty, risk or uncertainty – a particular decision rule is adopted for selecting the best plan(s) of action. We shall refer to this process as *decision-tree (DT) deliberation*.

The main thrust of this paper is to show how decision-tree deliberation can be utilized within a framework that is suited to resource-bounded reasoning in dynamic domains. We first describe the possible-worlds model and the decision tree representation formally. We then provide a transformation from decision trees to the possible-worlds model. From the possible worlds viewpoint, this provides a concrete method for obtaining the probability and payoff distribution on the worlds. From a decision theory viewpoint, the transformation facilitates symbolic manipulation of decision-theoretic entities. Finally, we describe classical decision-tree deliberation and show how it can be used to determine the formation of intentions.

## 3 POSSIBLE WORLDS MODEL

In our earlier work [15] we extended the propositional branching-time logic CTL* [4] to a possible-worlds framework and introduced modal operators for beliefs, goals, and intentions. In this section, we enhance this logic by introducing operators for probability (similar to that of Fagin and Halpern [5]) and payoffs.

### 3.1 SYNTAX AND SEMANTICS

Similar to CTL*, we have two types of formulas in our logic: *state formulas* (which are true in a specific world at a particular time point) and *path formulas* (which are true along a specific path). A state formula is defined as follows: (a) any propositional formula is a state formula; (b) if $\phi_1, ..., \phi_k$ are state formulas, $\psi_1, ..., \psi_k$ are path formulas, and $\theta_1, ..., \theta_k, \alpha$ are real numbers, then $\theta_1 \mathsf{PROB}(\phi_1) + ... + \theta_k \mathsf{PROB}(\phi_k) \geq \alpha$ and $\theta_1 \mathsf{PAYOFF}(\psi_1) + ... + \theta_k \mathsf{PAYOFF}(\psi_k) \geq \alpha$ are also state formulas; (c) if $\phi_1$ and $\phi_2$ are state formulas, and $\psi$ is a path formula, then $\neg\phi_1$, $\phi_1 \vee \phi_2$, $\mathsf{BEL}(\phi_1)$, $\mathsf{GOAL}(\phi_1)$, $\mathsf{INTEND}(\phi_1)$ and $\mathsf{OPTIONAL}(\psi)$ are state formulas. A path formula can be defined as follows: (a) any state formula is a path formula; (b) if $e$ is an event type then $done(e)$ is a path formula; (c) if $\psi_1$ and $\psi_2$ are path formulas, then $\neg\psi_1$, $\psi_1 \vee \psi_2$, and $\Diamond\psi_1$, are path formulas. Event types include primitive event types, $e_1;e_2$, and $?\phi$.

We now define formally the notion of an interpretation in our language.

**Definition 1** : An interpretation $M = <W, E, T, \prec, \mathcal{B}, \mathcal{G}, \mathcal{I}, \mathsf{PA}, \mathsf{OA}, \Phi>$. W is a set of worlds, E is a set of primitive event types, T is a set of time points, $\prec$ a binary relation on time points,[2] and $\Phi$ is a truth assignment of primitive propositions for any given world and time point. A situation is a world, say $w$, at a particular time point, say $t$, and is denoted by $w_t$. The relations, $\mathcal{B}$, $\mathcal{G}$, and $\mathcal{I}$ map the agent's current situation to her belief, goal, and intention-accessible worlds, respectively. More formally, $\mathcal{B} \subseteq W \times T \times W$ and similarly for $\mathcal{G}$ and $\mathcal{I}$. PA is a probability assignment function that assigns to each time point $t$ and world $w$ a probability function $\mu_t^w$. Each $\mu_t^w$ is a discrete probability function on the set of worlds W. OA is a payoff assignment function that assigns to each time point $t$ and world $w$ a payoff function $\rho_t^w$. Each $\rho_t^w$ is a partial mapping from paths to real-valued numbers.

**Definition 2** : Each world $w$ of W is a tuple $<T_w, \mathcal{A}_w, \mathcal{O}_w>$, where $T_w \subseteq T$, $\mathcal{A}_w \subseteq T_w \times T_w$, and $\mathcal{O}_w: T_w \times T_w \mapsto E$. Intuitively, $\mathcal{O}_w$ is an arc function that is a partial mapping from time points to an event and signifies the occurrence of an event. Also, $\mathcal{A}_w$ obeys the ordering of $\prec$. Such worlds are called *time trees*. A *fullpath* in a world $w$ is an infinite sequence of time points $(t_0, t_1, ...)$ such that $\forall i \, (t_i, t_{i+1}) \in \mathcal{A}_w$. We use the notation $(w_{t_0}, w_{t_1}, ...)$ to make the world of a particular fullpath explicit.

The semantics of the language with interpretation $M$ is as follows:

---

[2] We require that the binary relation be total, transitive and backward-linear to enforce a single past and branching future.

$M, w_{t_0} \models \mathsf{PROB}(\phi) \geq \alpha$ iff
   $\mu_{t_0}^w(\{w' \in \mathcal{B}_{t_0}^w \mid M, w'_{t_0} \models \phi\}) \geq \alpha$
$M, w_{t_0} \models \mathsf{PAYOFF}(\psi) \geq \alpha$ iff
   $\forall w' \in \mathcal{G}_t^w$ and $\forall x_i$ such that $M, x_i \models \psi$,
   where $x_i$ is a fullpath $(w'_{t_0}, w'_{t_{i1}}, ...)$,
   it is the case that $\rho_{t_0}^w(x_i) \geq \alpha$
$M, w_{t_0} \models \mathsf{OPTIONAL}(\psi)$ iff
   there exists a fullpath in $w$, $(w_{t_0}, w_{t_1}, ...)$
   such that $M, (w_{t_0}, w_{t_1}, ...) \models \psi$
$M, w_{t_0} \models R(\phi)$ iff $\forall \, w' \in \mathcal{R}_{t_0}^w$, we have $M, w'_{t_0} \models \phi$.
$M, (w_{t_0}, w_{t_1}, ...) \models \phi$ iff $M, w_{t_0} \models \phi$.
$M, (w_{t_0}, w_{t_1}, ...) \models \Diamond\phi$ iff
   $\exists k, k>0$ such that $M, (w_{t_k}, ...) \models \phi$
$M, (w_{t_1}, ...) \models done(e)$ iff
   there exists $t_0$ such that $e \in \mathcal{O}_w(t_0, t_1)$
$M, (w_{t_1}, ...) \models done(e_1;e_2)$ iff
   there exists $t_0$ such that $e_2 \in \mathcal{O}_w(t_0, t_1)$ and
   $M, (w_{t_0}, ...) \models done(e_1)$
$M, (w_{t_1}, ...) \models done(?\phi)$ iff $M, w_{t_1} \models \phi$.

R and $\mathcal{R}$ indicate the modal operators and relations, respectively, of belief, goal, and intention. We use the abbreviation $\mathcal{R}_t^w$ to denote all the worlds $\mathcal{R}$-accessible from $w$ at $t$.

The semantics of temporal and modal operators is relatively straightforward. The probability of a formula $\phi$ is greater than or equal to $\alpha$ if and only if the probability distribution of all the belief-accessible worlds in which $\phi$ is true is greater than $\alpha$. The payoff of a formula $\psi$ is greater than or equal to $\alpha$ if and only if the payoff function assigns a value greater than or equal to $\alpha$ to all paths where $\psi$ is true in all goal-accessible worlds.

$\mathsf{INEVITABLE}(\phi)$ is defined as $\neg\mathsf{OPTIONAL}(\neg\phi)$; $\Box\phi$ as $\neg\Diamond\neg\phi$. Additionally, the conditional probability $\mathsf{PROB}(\phi_1 \mid \phi_2) \geq \alpha$ can be represented as $\mathsf{PROB}(\phi_1 \wedge \phi_2) \geq \alpha.\mathsf{PROB}(\phi_2)$ [8].

We shall illustrate the belief- and goal-accessible worlds of an agent using a simple example. Phil, who is currently in the House of Representatives, believes that he can stand for the House of representatives (Rep), switch to the Senate and stand for a Senate seat (Sen), or retire from politics (Ret) [10]. He does not consider the option of retiring seriously and is sure to retain his House seat. He has to make a decision regarding conducting or not conducting an opinion poll, based upon which he has to decide to stand for the House or the Senate. The results of the poll would be either a majority approving his switch to the Senate (yes) or a majority disapproving of his switch (no).

Consider the current situation to be $w_t$. The four belief-accessible worlds of $w_t$, shown in Figure 1, correspond to Phil winning or losing the Senate seat based on the majority answering yes or no in the poll. The probabilities of these worlds are shown in the top right hand corner of each world. The propositions *win*, *loss*, *yes*, and *no* are true at the situations shown.

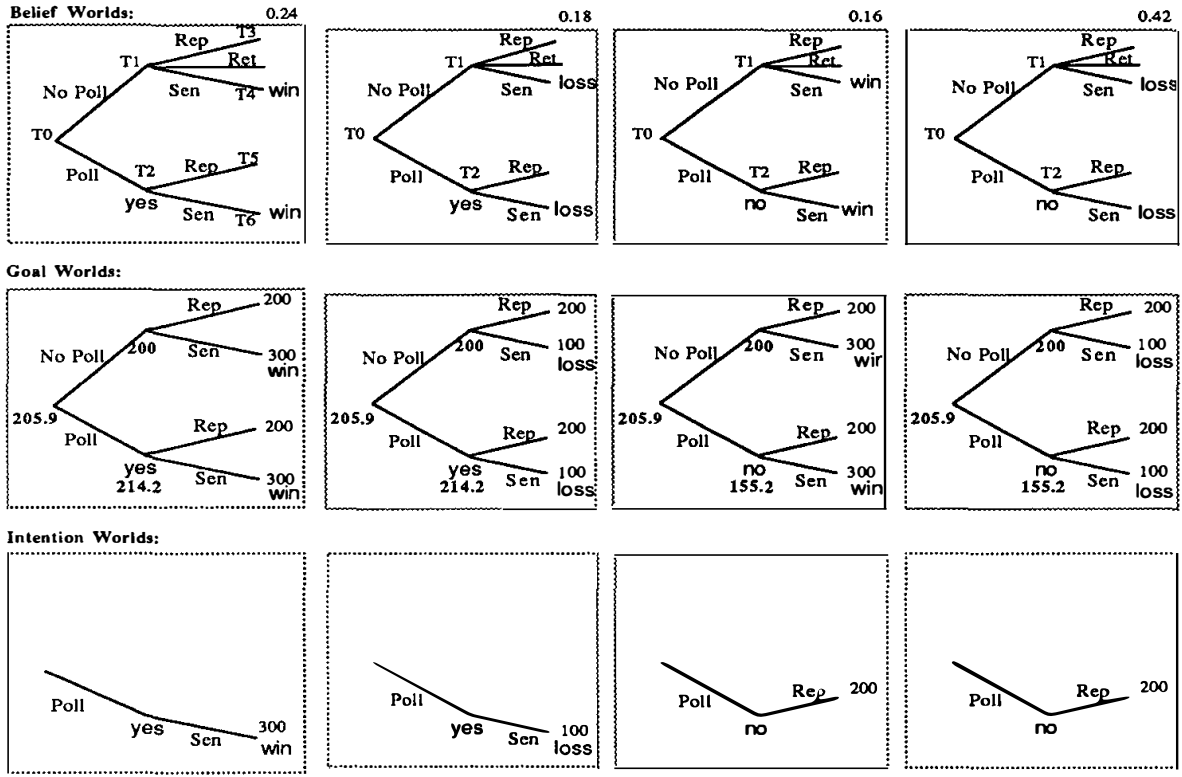

Figure 1: Belief, Goal, and Intention (wrt *maxexpval*) Worlds

Some of the formulas that are satisfiable at $w_t$ are BEL(OPTIONAL($\Diamond done(Sen)$)), i.e., Phil believes that he has the option of eventually standing for the Senate, and PROB(OPTIONAL($\Diamond yes$)) = 0.42, i.e., the probability of eventually achieving a yes response is 0.42.

The goal worlds are also shown in Figure 1 (for clarity, we have omitted the time points, which are the same as in the belief worlds). The values at the end of the paths (100, 200, and 300) signify the value of losing a Senate seat, winning a House seat, and winning a Senate seat. This can be expressed as PAYOFF($\Diamond(done(Sen)$ $\wedge loss$)) = 100, PAYOFF($\Diamond(done(Sen) \wedge win)$) = 300, etc. Other formulas can state other properties of the agent's goals. For example, the goal of the agent to retain his option to eventually stand for a Senate seat is expressed as GOAL(OPTIONAL($\Diamond done(Sen)$)). Note that the option of retiring from politics exists only in belief worlds, not in goal worlds, i.e., Phil believes that retiring is an option, but does not have any goal towards retiring.

### 3.2 SEMANTIC CONDITIONS

In this section, we give an informal description of some of the semantic conditions that can be imposed on our possible worlds model. Some of these conditions require the definition of a *sub-world*. We define a world to be a *sub-world* of another (denoted by $\sqsubseteq$) if and only if the time points in one are a subset of the other, they share the same history, and everything else is identical. The formal definition of sub-worlds and the axioms corresponding to the following semantic conditions are given elsewhere [13].

If the belief-accessible worlds represent chance, then no level of introspection can change the chance. Thus we require that all belief-accessible worlds have identical probability distributions (Semantic Condition C1). We also require that the probability distribution over belief-accessible worlds add up to one (C2). From C2 we also have that beliefs about inevitable facts have probability one.

We introduce a constraint on our belief, goal, and intention-accessible worlds, called *strong realism* [15]. Strong realism requires that for every belief-accessible world there exists a sub-world which is a goal-accessible world, and for every goal-accessible world there exists a sub-world which is an intention-accessible world (C3). The same restriction can be applied in the reverse direction also (C4). These two conditions essentially ensure that, if the agent intends an option, he has the goal towards that option and also believes in that option. These semantic conditions are very strong and have a significant impact on the inter-relationships between beliefs, goals, and intentions. We show elsewhere [14, 15] how these conditions can be relaxed to solve some of the problems associated with possible-world representations of beliefs, goals, and intentions [3].

More formally, the semantic conditions C1 to C4 can be stated as follows:

(C1) $\forall w' \in \mathcal{B}_t^w, \mu_t^w = \mu_t^{w'}$.
(C2) $\mu_t^w(\mathcal{B}_t^w) = 1$.
(C3) $\forall w' \in \mathcal{B}_t^w \; \exists w'' \in \mathcal{G}_t^w$ such that $w'' \sqsubseteq w'$ and
$\forall w' \in \mathcal{G}_t^w \; \exists w'' \in \mathcal{B}_t^w$ such that $w'' \sqsubseteq w'$.
(C4) $\forall w' \in \mathcal{G}_t^w \; \exists w'' \in \mathcal{I}_t^w$ such that $w'' \sqsubseteq w'$ and
$\forall w' \in \mathcal{I}_t^w \; \exists w'' \in \mathcal{G}_t^w$ such that $w'' \sqsubseteq w'$.

In the remainder of this paper, we shall use this possible-worlds BDI model as a basis for deliberation.

## 4 DECISION TREES AND GOAL WORLDS

In this section, we give a formal description of a decision tree and show how one can transform a decision tree into a set of goal-accessible worlds. Intuitively, both decision trees and goal-accessible worlds capture the desirable ends or outcomes of the decision problem, the different alternatives or choices available to the agent to achieve those ends, and the chance events controlled by nature. Note that this intuitive mapping is possible only because we have chosen to represent each possible world as a branching-time structure, rather than the more traditional model where each possible world is a linear-time structure [3]. Although one may be able to define a transformation from decision trees to linear-time models, we believe that such a mapping would be less intuitive than the one illustrated here.

The decision tree for our running example is given on the left hand side of Figure 3. Decision nodes are denoted by boxes and chance nodes by circles. The formal definition of a decision tree is as follows:

**Definition 3** : A *decision tree* DT = $<\mathcal{N}, \mathcal{E}, \mathcal{S}, \mathcal{PS}, \mathcal{P}, \mathcal{U}, \Phi, \Psi, \Sigma>$. $\mathcal{N}$, is the union of all decision nodes $\mathcal{D}$, all chance nodes $\mathcal{C}$, and all terminal nodes $\mathcal{T}$. $\Phi$ is the set of all propositional formulas, $\Psi$ is the set of all probabilistic state formulas (which includes the conditional probability operator in addition to the standard logical operators), and $\Sigma$ is the set of all primitive event types. $\mathcal{E} \subseteq \mathcal{D} \times \mathcal{N} \times \Sigma$ is an event relation. $\mathcal{S} \subseteq \mathcal{C} \times (\mathcal{D} \cup \mathcal{T}) \times \Phi$ is a chance relation. $\mathcal{PS} \subseteq \mathcal{C} \times (\mathcal{D} \cup \mathcal{T}) \times \Psi$ is a probabilistic state relation. $\mathcal{P}: \Psi \to \Re$ is a probability function that maps probabilistic states to real numbers. $\mathcal{U}: \mathcal{T} \to \Re$ is the payoff function that assigns to a terminal node a real number.

Now we consider the transformation from decision trees to possible worlds. Given a decision tree, we start from the root node and traverse each arc. For each unique state labeled on an arc emanating from a chance node,[3] we create a new decision tree that

---
[3]Note that the decision tree is split with respect to the

```
create(t, n, p)
    Case n is a decision node
        For all m such that ε(n, m, e)
            create(t, m, p);
    Case n is a chance node
        For all s such that S(n, m, s)
            For all (u q) in remove(t, s, p)
                create(u, m, q);
    Case n is a terminal node
        return(t, p).
remove(t, s, p)
    For all n, m such that S(n, m, s) and PS(n, m, r)
        collect(
            (t - S(n, m, s) - ε(k, n, e) + ε(k, m, e)), p.r))
```

Figure 2: Functions for Transformation

is identical to the original tree except that (a) the chance node is removed and (b) the arc incident on the chance node is connected to the successor of the chance node. This process is carried out recursively until there are no chance nodes left. Each of the decision trees so obtained consists of only decision nodes and terminal nodes. Each one of these decision trees is then transformed into a possible world structure by appropriately renaming the relations. The payoff function is assigned to paths in a straightforward way, thus yielding a set of goal-accessible worlds.

We obtain the probability distribution over the corresponding belief worlds by associating with each decision tree that is created a value $\alpha$, which will finally correspond to the probability of a goal world. This probability is essentially the weighted product of all the chance nodes that a particular world represents. This probability distribution is finally passed back onto the corresponding belief-accessible worlds.

The transformation is performed by two functions, **create** and **remove**, which are defined in Figure 2. We have assumed in the function **remove** that the chance node is connected by an arc from a decision node. This is true in all cases except when the chance node is the root node of the decision tree. We have also assumed that the chance states are named uniquely.

The **create** function, when called with a given decision tree, its root node, and a probability value of one, will result in a set of decision trees with appropriate probabilities. The final transformation from these multiple decision trees with no chance nodes to possible worlds is trivial and is given elsewhere [13]. Figure 3 gives the transformation for the running example.

The possible worlds so formed are goal-accessible worlds. The probabilities associated with these worlds are the same as the probabilities of the decision trees

---
chance states and not with respect to the chance nodes. This is important to avoid invalid goal worlds.

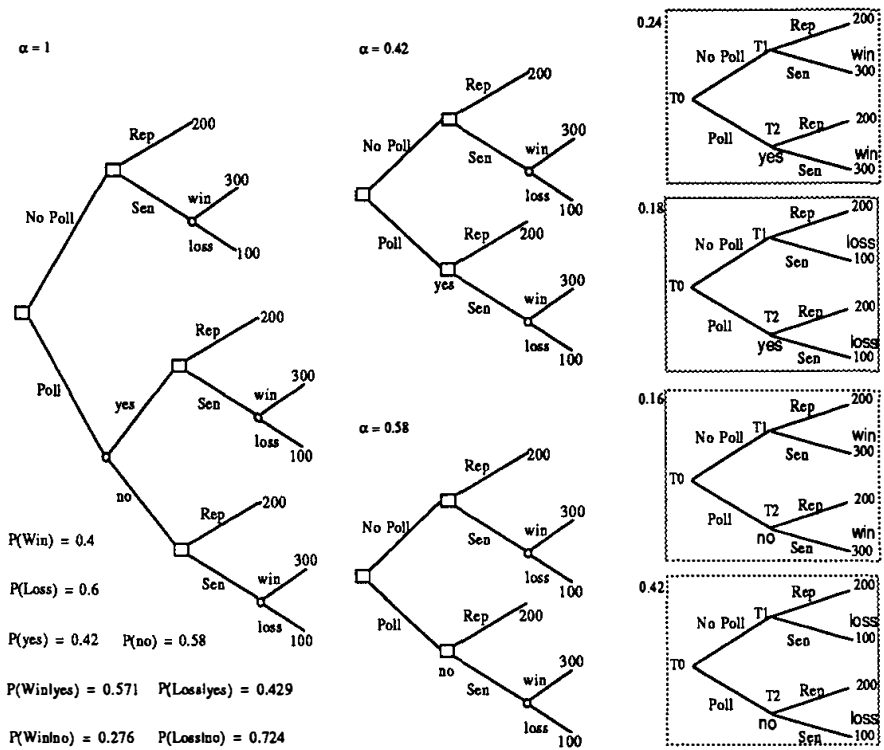

Figure 3: Transformation from Decision Trees to Goal Worlds

from which they are derived. Given our semantic condition earlier that all goal-accessible worlds have corresponding belief-accessible worlds, the probability distribution flows backwards to belief-accessible worlds. This transformation yields the following proposition:

**Proposition 1** : *Given a decision tree DT we can create a possible worlds interpretation M such that the information given by the decision tree DT is satisfiable for a particular world y and time t in M. We shall denote this by* transform($DT,<M,y,t>$).

## 5 DELIBERATION AND INTENTIONS

Given a decision tree and the above transformation, an agent can make use of standard decision-theoretic techniques such as *maximin* or *maximizing expected value* to deliberate and decide the best plan of action. This best plan of action is what the agent *commits* to and adopts as an intention.

To capture the process of decision theory deliberation, we introduce two generic functions, the *value function*, denoted by $\mathcal{V}$, and the *deliberation function*, denoted by $\delta$. The value function assigns a real-valued number to every node in the decision tree and the deliberation function chooses one or more best sequences of actions to perform at a given node. Both these functions will be parameterised on the particular deliberation procedure used; i.e., maximin deliberation, maximizing the expected utility, or any other deliberation procedure. We shall use the operator ';' to denote sequencing of actions.

First we consider the maximin approach. The value and deliberation functions for this approach is given in Figure 4. For the running example, the maximin deliberation function returns the set $\{No\ Poll;Rep,\ Poll;Rep\}$.

Next we examine the principle of maximizing expected value. For decision nodes and terminal nodes, the value function and the deliberation function using the maxexpval principle are identical to the ones under the maximin principle. For chance nodes, the value and deliberation functions are defined as:

$\mathcal{V}(maxexpval,\ n_i) = \sum_{\{n_j|\mathcal{PS}(n_i,n_j,p_j)\}} P(p_j).\mathcal{V}(n_j)$

$\delta(maxexpval,\ n_i) =$
$\quad \{s_j?; \delta(maxexpval, n_j)|\mathcal{S}(n_i, n_j, s_j)\}$

For the running example the above deliberation function returns $\{Poll;yes?;Sen,\ Poll;no?;Rep\}$. Actions *yes?* and *no?* are used as conditional tests.

Modifying the modal operator INTEND with a subscript that indicates the decision procedure used, we can state the following theorem which allows an agent to form intentions based on his deliberation.

**Theorem 1** : *If an agent with a decision tree DT with*

$$\mathcal{V}(maximin, n_i) = \begin{cases} \min_{\{n_j | \mathcal{S}(n_i, n_j, s_j)\}} \mathcal{V}(n_j) & \text{if } n_i \in \mathcal{C} \\ \max_{\{n_j | \mathcal{E}(n_i, n_j, e_j)\}} \mathcal{V}(n_j) & \text{if } n_i \in \mathcal{D} \\ \mathcal{U}(n_i) & \text{if } n_i \in \mathcal{T} \end{cases}$$

$$\delta(maximin, n_i) = \begin{cases} \{\delta(maximin, n_j) \mid \mathcal{S}(n_i, n_j, s_j) \text{ and } \mathcal{V}(n_j) = \mathcal{V}(n_i)\} & \text{if } n_i \in \mathcal{C} \\ \{e_j; \delta(maximin, n_j) \mid \mathcal{E}(n_i, n_j, e_j) \text{ and } \mathcal{V}(n_j) = \mathcal{V}(n_i)\} & \text{if } n_i \in \mathcal{D} \\ nil & \text{if } n_i \in \mathcal{T} \end{cases}$$

Figure 4: Value and deliberation functions for *maximin*

root node $n$ and deliberation procedure $d$ chooses as a best plan a sequence of actions $a$, i.e., $a \in \delta(d, n)$, and the transformation is transform($DT$, $<M,y,t>$) then $M, y, t \models \text{INTEND}_d(\text{OPTIONAL}(\diamond done(a)))$.

To prove the above theorem, we need to define the process of deliberation within a possible-worlds model that generates intention-accessible worlds from a given set of goal-accessible and belief-accessible worlds. This definition is given elsewhere [13]. The set of intention-accessible worlds generated by maxexpval deliberation is shown in Figure 1. The possible-worlds deliberation can be shown to be equivalent to the decision tree deliberation [13]. This equivalence together with Proposition 1 establishes the above theorem.

Note that the maximin deliberation function always commits to a particular branch emanating from a chance node; namely, the branch that leads to a minimum value node. This means that $\delta$ for *maximin* can always return a sequence of actions without being conditional on any state information. However, this is not true in the case of *maxexpval* deliberation because each chance node is a weighted sum of all its branches. The actions thus have to be conditional on the state of the world. Thus *maximin* deliberation yields unconditional intentions and *maxexpval* deliberation results in conditional intentions.

For the example under discussion, we have the following unconditional maximin intentions and conditional maxexpval intentions: (a) the agent intends by maximin that, in all future paths he will stand for the House of representatives, i.e., $\text{INTEND}_{maximin}$ (INEVITABLE ($\diamond done(Rep)$)); (b) the agent intends by maxexpval that, in all future paths in which he has carried out a poll and the majority have answered yes, he would stand for the Senate, i.e., $\text{INTEND}_{maxexpval}$ (INEVITABLE($\diamond$ ($done(Poll) \wedge yes \supset \diamond done(Sen)$))); and (c) the agent intends by maxexpval that, in all future paths in which he has carried out a poll and the majority have answered no, he would stand for the House, i.e., $\text{INTEND}_{maxexpval}$ (INEVITABLE($\diamond$ ($done(Poll) \wedge no \supset \diamond done(Rep)$))).

So far, we have discussed how deliberation leads to the formation of conditional and unconditional intentions. As discussed in the introduction, intentions play the two important roles of posing decision problems for deliberation and constraining the options open for deliberation. The decision tree and the decision problem that we until now have taken for granted could have been generated because of a top-level intention of Phil to be rich and famous. In other words, a prior intention of the form $\text{INTEND}(\text{OPTIONAL}\diamond(rich \wedge famous))$ followed by means-end reasoning could have resulted in the decision tree to conduct a poll and run for the House or Senate seat. Also, if Phil had the prior intention to stand for a Senate seat, i.e., if $\text{INTEND}(\text{INEVITABLE}\diamond done(Sen))$ were true, then the decision tree would be one without any alternatives for standing for the House. Thus, future-directed intentions constrain the decision problem that has to be considered.

We believe that the formalism we have presented here is general enough to cover a wide range of decision problems. The transformation and equivalence established in this paper should help one to choose the appropriate representation for the appropriate purpose, making use of the results of one representation within the other. For example, one could operate within a possible-worlds BDI framework for reasoning about the interaction of beliefs, goals, and intentions, and how they change with time [15], shift to a decision tree representation for deliberation, and then come back to the possible-worlds framework for reasoning about the intentions so formed.

## 6   CONCLUSIONS

This paper examines one of the important philosophical aspects of Bratman's theory of rational agency; namely, that deliberation leads to the formation of intentions. We have presented a powerful branching-time possible-worlds model for reasoning about beliefs, goals, intentions, actions, time, probabilities, and payoffs, and provided a transformation from decision trees to structures in this model. We have also shown how the deliberation procedure used determines the intentions adopted by the agent. This formal model of deliberation within a BDI-architecture is one of the main contributions of this paper. Previous work on formalizations of BDI-architectures [3, 12, 15] does not ad-dress this issue.

Recent work in real-time reasoning has vigorously pursued the use of decision-theoretic techniques. Russell and Wefald [16] treat computations themselves as actions, with appropriate utilities. These computations have to be chosen from among a number of different alternatives and decision theory is used to choose the best action or computation. This facilitates meta-level reasoning. Haddawy and Hanks [9] explore the relationships between symbolic goals and numeric utilities. In particular, they address the problem of building utility functions. However, neither approach considers the role of decision theory in the formation of intentions, which is the primary focus of this paper.

Fagin and Halpern [5] combine reasoning about knowledge and probabilities by explicitly introducing probability formulas. Haddawy [8] introduces reasoning about probabilities in a branching time model by considering a world to be a future path. The work presented here deals with both future paths in a branching time model and different possible worlds in the epistemic sense. It also introduces explicit reasoning about payoffs and treats payoffs as values the agent places on his future paths within a goal-accessible world. Thus it builds on the existing tradition by combining a possible-worlds BDI framework with decision-theoretic deliberation and explicit reasoning about probabilities and payoffs.

Our future work in this area will focus on the role of intentions in deliberation and reconsideration. We aim to analyze the need for rational agents to reconsider intentions and when they should carry out such reconsideration.